\pdfoutput=1
\documentclass[10pt,twocolumn,letterpaper]{article}

\usepackage[pagenumbers]{cvpr} 

\usepackage{graphicx}
\usepackage{amsmath}
\usepackage{amssymb}
\usepackage{booktabs}
\usepackage[pagebackref,breaklinks,colorlinks]{hyperref}

\usepackage{latexsym}
\usepackage{booktabs}
\usepackage{algorithm}
\usepackage{algorithmic}
\usepackage[switch]{lineno}

\usepackage{multirow}
\usepackage{makecell}
\usepackage{pifont}
\usepackage{colortbl}
\usepackage{xcolor}
\usepackage{subcaption}
\usepackage{diagbox}
\usepackage{bm}

\newcommand{\suptext}[1]{$^{\text{#1}}$}

\definecolor{demphcolor1}{gray}{.6}
\definecolor{sherrybrown}{RGB}{227,207,87}
\definecolor{lightred}{RGB}{241,140,142}
\definecolor{deepgreen}{RGB}{127, 161, 104}
\definecolor{lightblue}{RGB}{122,46,246}

\newcommand{\demphs}[1]{\textcolor{demphcolor1}{#1}}

\usepackage[capitalize]{cleveref}
\crefname{section}{Sec.}{Secs.}
\Crefname{section}{Section}{Sections}
\Crefname{table}{Table}{Tables}
\crefname{table}{Tab.}{Tabs.}

\def\cvprPaperID{*****} 
\def\confName{CVPR}
\def\confYear{2022}

\begin{document}

\title{Bridging Generative and Discriminative Models for Unified Visual Perception with Diffusion Priors}

\author{Shiyin Dong\suptext{1}
\and
Mingrui Zhu\suptext{1}\thanks{Corresponding author}
\and
Kun Cheng\suptext{1} 
\and
Nannan Wang\suptext{1} 
\and
Xinbo Gao\suptext{2} 
\and
\suptext{1}Xidian University\\
\suptext{2}Chongqing University of Posts and Telecommunications\\
}
\maketitle

\begin{abstract}
    The remarkable prowess of diffusion models in image generation has spurred efforts to extend their application beyond generative tasks.
    However, a persistent challenge exists in lacking a unified approach to apply diffusion models to visual perception tasks with diverse semantic granularity requirements. 
    Our purpose is to establish a unified visual perception framework, capitalizing on the potential synergies between generative and discriminative models. In this paper, we propose Vermouth\footnote{Our model is named after vermouth. Indeed, the properties of diffusion and blending of wines and herbs in vermouth share similarities with our design.}, a simple yet effective framework comprising a pre-trained Stable Diffusion (SD) model containing rich generative priors, a unified head (U-head) capable of integrating hierarchical representations, and an adapted expert providing discriminative priors.  Comprehensive investigations unveil potential characteristics of Vermouth, such as varying granularity of perception concealed in latent variables at distinct time steps and various U-net stages.
    We emphasize that there is no necessity for incorporating a heavyweight or intricate decoder to transform diffusion models into potent representation learners.
    Extensive comparative evaluations against tailored discriminative models showcase the efficacy of our approach on zero-shot sketch-based image retrieval (ZS-SBIR), few-shot classification, and open-vocabulary semantic segmentation tasks.
    The promising results demonstrate the potential of diffusion models as formidable learners, establishing their significance in furnishing informative and robust visual representations.
    
\end{abstract}

\section{Introduction}


\begin{figure}[t]
\setlength{\abovecaptionskip}{0.2cm}
    \centering
    \resizebox{0.8 \linewidth}{!}{
    \includegraphics[width= \linewidth]{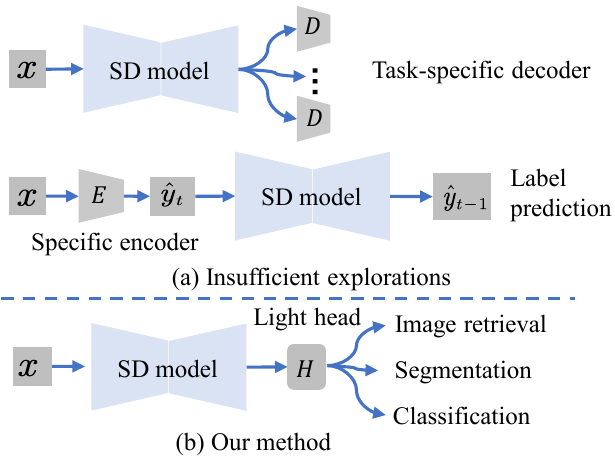}
    }
    \caption{We transfer priors of the SD model in a unified framework for different visual perception tasks.}
    \label{fig:teaser}
\end{figure}

\begin{quotation}
    ``\emph{What I cannot create, I do not understand.}"
\end{quotation}

    \hfill ---Richard P. Feynman, 1988
    
    Over the years, there has been a keen interest in exploring the representation learning capabilities of generative models, given their proficiency in generating vivid images. Consistent with Feynman's perspective, to produce high-fidelity samples, a generative model must attain a high level of semantic understanding. Early works~\cite{DAE,denoising-autoencoders} have demonstrated  that generative models can be effectively employed for discriminative tasks through non-trivial methods.

    Recently, diffusion models have emerged with stunning performance in the image generation field, creating realistic and incredibly detailed images~\cite{dhariwal:beat-gans,ho:classifier-free-guidance,ho:ddpm}. Specifically, large-scale text-to-image diffusion models~\cite{rombach:stable-diffusion,parmar2023zero} can seamlessly integrate and modify semantic information in an end-to-end manner, which allows the synthesis of images featuring diverse objects, scenes, and styles~\cite{zhang:control-net,instructpix2pix}. Given this phenomenon, we deem that large text-to-image diffusion models, such as SD~\cite{rombach:stable-diffusion} model, have acquired high-level and low-level semantic cues through extensive exposure to large-scale image-text pairs. Nevertheless, the question of how to extract the latent knowledge embedded in the diffusion process and harness this knowledge for visual perception tasks remains an unsolved challenge.

    Visual perception tasks necessitate the establishment of distinct decision boundaries $p_{\theta}(y|x)$ among various categories, an objective not initially envisioned in the design of diffusion models. This achievement is typically attained through supervised learning~\cite{swin-v1,convneXt-v1}, unsupervised contrastive learning~\cite{dino-v1}, and masked image modeling followed by supervised fine-tuning~\cite{he2022mae,bao2021beit-v1}. In contrast, diffusion models aspire to model the inherent probability distributions $p_{\theta}(x|z)$ found in a dataset. Consequently, an inherent incompatibility exists between diffusion models and visual perception tasks. As shown in Figure~\ref{fig:teaser}, various methods~\cite{zhao2023:VPD,xu2023:ODISE,karazija2023:OVDiff} have attempted to address this issue by integrating off-the-shelf and heavy decoders or encoders with SD models, yielding favorable outcomes. Nevertheless, the exploration of how to effectively leverage hierarchical features within SD models in a unified framework and the factors (e.g., time steps) influencing such features are inadequate.

    The goal of this paper is to scrutinize the release of internal priors within diffusion models and their transfer to non-generative tasks in a unified manner. 
    In contrast to approaches involving combined tuning with various decoders, we propose a simple yet effective method applicable to tasks with diverse granularity requirements within a unified framework.
    Specifically, for the input image, we introduce suitable noise and project it into the latent space of the SD model under accurate text guidance. Subsequently, a unified head (U-head) blends latent representations from different granularities, eliminating the need for designing complex and tailored decoders. Moreover, this architecture demonstrates remarkable flexibility, enabling smooth integration with the priors of discriminative models, denoted as adapted expert, resulting in enhanced compatibility with visual perception tasks. Leveraging the flexibility and effectiveness of this module and the rich semantic features embedded within diffusion, we can seamlessly transfer the fused features to diverse tasks.
    
    To comprehensively analyze the most effective way for unlocking the knowledge within the SD model, we investigated three downstream tasks across a range of near-real-world scenes, including ZS-SBIR, open-vocabulary semantic segmentation, and few-shot classification tasks to evaluate our method. Experiment results on over 20+ datasets reveal that, despite its inherent mismatch with visual perception tasks, the SD model can still be regarded as a promising learner.

    In summary, our contributions are as follows:
    \begin{itemize}
        \item To the best of our knowledge, we are the first to propose a unified framework to apply diffusion to visual perception tasks demanding different granularity semantics.
        \item We design a unified head capable of effectively fusing the generative priors of SD models with discriminative priors from the adapted expert.
        \item Comprehensive experiments and analyses of diffusion features unveil observed rules for hyperparameters such as the noise level of latents, which can provide constructive insights and suggestions for subsequent research.
    \end{itemize}

\begin{figure*}[t]
        \setlength{\abovecaptionskip}{0.1cm}
        \centering
        \includegraphics[width= \linewidth]{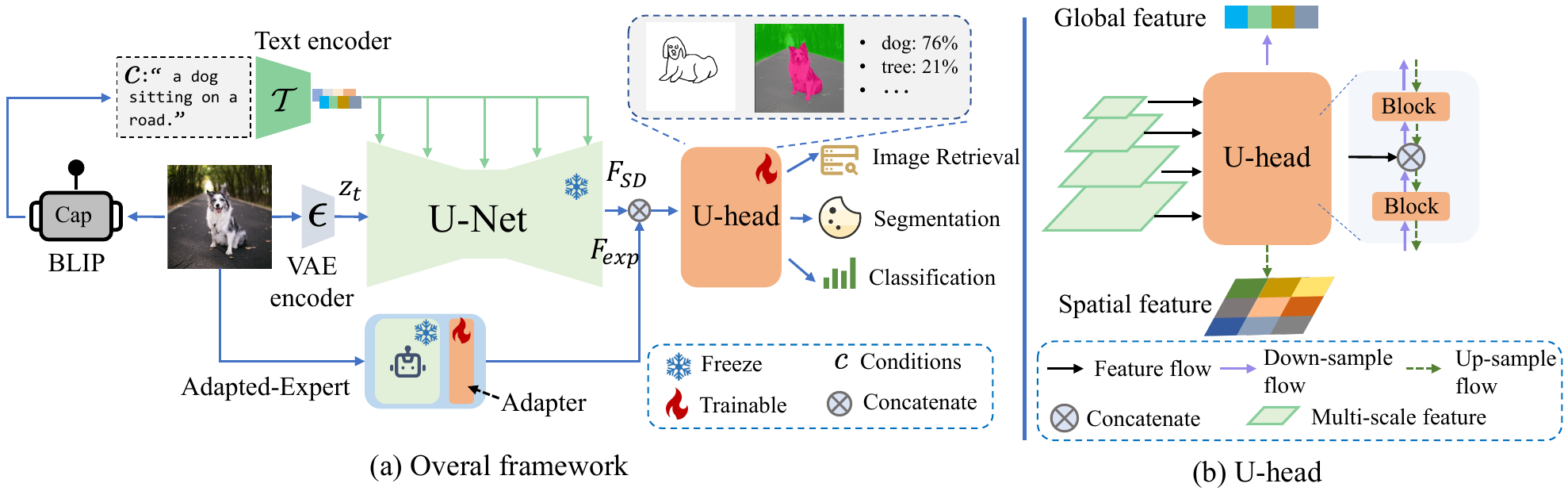}
        \caption{An overview of our framework. We employ the pre-trained BLIP model to acquire an accurate description and utilize the SD model to generate representations guided by text embedding. The introduction of the U-head is intended to fuse the two representations from the SD model and the adapted expert, aiming to enhance compatibility in discriminative tasks with different semantic granularity requirements.}
        \label{fig: framework}
    \end{figure*}

\section{Related Work}
\subsection{Vision Models in Perception Task}

    The paradigm of pre-training and transfer learning has significantly advanced the domain of computer vision. In the early years, convolutional neural networks~\cite{resnet} and Vision Transformer (ViT)~\cite{dosovitskiy2020ViT} have been viewed as standard architectures for various vision tasks, owing to their exceptional global perception capabilities and adaptable scalability. Nevertheless, the conventional pre-training on fixed-label datasets constrains their applicability in some complex scenarios.

    Recently, CLIP~\cite{Radford:CLIP} has garnered significant attention across various domains due to its efficient visual-language alignment. Some approaches leverage CLIP for tasks such as few-shot classification\cite{gao2023clip-adapter}, image retrieval\cite{saito2023pic2word}, visual grounding~\cite{rao2022denseclip}, and image segmentation\cite{zhou2022maskclip}, which demonstrates the remarkable and rapid learning capabilities. In this paper, our focus shifts to the SD model, aiming to investigate whether it exhibits similar characteristics and how to harness its potential capabilities.

\subsection{Perceptual Learning with Diffusion Models}

    The application of generative models to discriminative tasks, exemplified by BigBiGAN~\cite{donahue2019bigbigan}, achieving a promising result on ImageNet, has sustained substantial interest over an extended period. Diffusion models~\cite{ho:ddpm}, exemplified by the SD model, have not only excelled in the field of image synthesis but have also drawn considerable attention in other domains. Harnessing latent features, the SD model exhibits versatility, finding applications in diverse domains, including classification~\cite{wei2023diffmae}, image segmentation~\cite{xu2023:ODISE,li2023grounded-diffusion}, depth estimation~\cite{ji2023ddp,zhao2023:VPD}, and semantic correspondence~\cite{zhang2023sd-dino}. 

    We focus on features-based methods mentioned above, which we categorize into two groups: specific encoder-based~\cite{ji2023ddp} and specific decoder-based~\cite{zhao2023:VPD,xu2023:ODISE} approaches. These methods necessitate task-specific design that makes them complex. In contrast, we sidestep this tailored procedure and advocate a unified framework capable of accommodating various scenarios.

    We noted that the methods most akin to ours are VPD~\cite{zhao2023:VPD} and Grounded-Diffusion~\cite{li2023grounded-diffusion}. VPD employs tailored decoders for distinct tasks and utilizes ambiguous prompts for text guidance. Grounded-Diffusion relies on an additional pre-trained grounding model and employs iterative denoising from pure Gaussian noise to get clean images. In comparison to VPD, our architecture is more lightweight, unified, and flexible. Compared with Grounded-Diffusion, our application scenarios are broader and do not necessitate additional model-assisted (Further detailed discussion can be seen in the supplementary material).

\section{Method}

    Aligned to directly apply the SD model in non-generative domains within a unified framework, this section introduces Vermouth. It is devised to transfer the prior knowledge of the SD model for achieving visual perception at different granularities, eliminating the need for task-specific designs. We will start with a review of the preliminaries of diffusion models in Section~\ref{sec:preliminaries}. Subsequently, our core ideas, including the U-head for obtaining the final representation, and the methodology for combining the adapted expert to enhance compatibility between the SD priors and discriminative tasks, are detailed in Section~\ref{sec:U-head} and Section~\ref{sec:adapted expert}.

\subsection{Preliminaries}\label{sec:preliminaries}

    Diffusion models constitute a category of likelihood-based models inspired by non-equilibrium thermodynamics. These models can characterize the data distribution $p(x)$ by learning to reverse the forward process, which incrementally introduces noise to the data. The forward diffusion process at $t$-th time-step can be modeled as Markov: $z_{t} \sim q(z_t|z_{t-1}) = \mathcal{N}(\sqrt{1-\beta _t}z_{t-1},(\beta _t)\boldsymbol{I})$,  where $\beta _t$ is associated with the noise schedule~\cite{ho:ddpm}. By employing a reparameterization trick, we can simplify this expression into a more manageable form:
    \begin{align}
        q\left(z_{t} \mid z_{0}\right) &=\mathcal{N}\left(z_{t} ; \sqrt{\bar{\alpha}_{t}} z_{0},\left(1-\bar{\alpha}_{t}\right) \boldsymbol{I}\right), \\
        \bar{\alpha}_{t} &=\prod_{s=1}^{t} \alpha_{s} = \prod_{s=1}^{t}(1-\beta _{s}).
    \end{align} 
    Generally, diffusion models introduce noise to inputs until $z_{T} \sim \mathcal{N}(\boldsymbol{0}, \boldsymbol{I})$ and samples iteratively by denoising the latent variables:
    \begin{equation}
        p(z_{0:T}) = p(z_T)\prod_{t=1}^T p(z_{t-1}|z_{t}).
    \end{equation}
    Through the proper reparameterization trick, one can predict the noise component $\epsilon$ by neural network $\epsilon_{\theta}(x_t;t)$ implemented by a U-net to learn how to reconstruct the input data:
    \begin{equation}
        \mathcal{L}_{\text{simple}} = \mathbb{E}_{z_t,\epsilon \sim \mathcal{N}(0,1)}\left[\lVert \epsilon - \epsilon_{\theta}(z_t;t,c) \rVert_2^2\right],
    \end{equation}
    where $c$ is an additional condition such as text prompts.

    The SD model employs a VQ-VAE~\cite{van2017vqvae} encoder for projecting the input into the latent space, denoted as $z_0 = \mathcal{E}(x)$. Subsequently, a decoder is utilized to reconstruct the input, expressed as $\hat{x} = \mathcal{D}(\hat{z_0})$. For latent modeling, the SD model adopts an asymmetric U-net, which is structured into three stages: down-sample, bottleneck, and up-sample, encompassing a total of 18 blocks. See the supplementary for the architectural details of the SD model.

\subsection{Latent Prior in SD Model}\label{sec:Latent prior of sd model}

    Given the presence of multi-scale features in the U-net, one can utilize specific blocks in the U-net to extract features.
    \begin{equation}
        F_{\text{SD}} = \text{U-net}(z_t, t, c),
    \end{equation}
    where $F_{\text{SD}}=\{f_i \in \mathbb{R}^{H_i \times W_i \times C_i}|i \sim \bm{B}\}$ and $\bm{B}$ is indexes of the specific blocks.

    Recent works~\cite{zhao2023:VPD,xu2023:ODISE,zhang2023sd-dino} have opted for different blocks, combining them with specific decoders to extract features for discriminative tasks. However, there remains a lack of clear insight regarding which blocks' output features are more semantically rich. This challenge stems from the non-trivial nature of selecting specific blocks among a total of 18. We opt for a more macro level to investigate the semantic features within the SD model. In essence, we employ the stages as the research granularity to amalgamate semantic information at various levels. 
    \begin{equation}
        f_i = \text{U-net}_{\texttt{stage\_i}} (z_t, t, c),
    \end{equation}
    where \texttt{stage\_i} represent the stages index.
    

    As a text-to-image model, the text prompt plays a crucial role in feature extraction as it serves as guidance for semantic synthesis. An intuitive approach involves using all class names $s$ in the dataset $\mathcal{D}$ to form the text context:
    \begin{equation}
        c = \text{concat}(\left[\mathcal{T}(s)|s \in \mathcal{D}\right]),
    \end{equation}
    where $\mathcal{T}$ is the text encoder of CLIP. 
    However, employing the same text context may lead to potential misalignment, especially considering that different inputs will contain different objects. In contrast,  we enhance alignment by utilizing BLIP~\cite{li2022blip} caption model to derive the image-aligned text prompt $s=\text{Cap}(x)$. 
    The multimodally pre-trained BLIP model, for a given image, excels in generating accurate descriptions while preserving semantics, encompassing both the object and its surroundings present in the image $s \in \mathcal{X} \subset \mathcal{D}$.

    
    In addition to the output features $f_i$ at each resolution, the attention maps of U-net typically encapsulate rich semantic information. Inspired by recent works~\cite{zhao2023:VPD} considering cross-attention map as an extracted diffusion prior, we are motivated to investigate the impact of the cross-attention map $A=\text{Softmax}(\frac{QK^T}{\sqrt{d}})$, where $Q$ and $K$ denote the image and text hidden states.  However, it is noteworthy that this approach did not consistently yield positive results, as detailed in our experimental findings.

\subsection{U-Head for Perception Tasks}\label{sec:U-head}

    Our U-head is designed to receive multi-scale features extracted from the SD model, facilitating the capture of language-aware visual features. As illustrated in Figure~\ref{fig: framework}, to obtain global features, this module progressively fuses high-resolution features containing detailed semantics to low-resolution features with more high-level semantics along the down-sample flow:
    \begin{equation}
        h = \text{U-head}(F),
    \end{equation}
    where $h$ is the final representation. Contrarily, detailed pixel-level features can be acquired along the up-sample flow. This approach encourages the capture of visual features at various levels of granularity, seamlessly blending coarser to finer semantic cues. Consequently, the need for specially customized redundant decoders\footnote{We use the terms decoder and head to distinguish modules of different magnitudes. In general, the decoder is larger than the head and has more parameters.} is mitigated to a certain extent. Then, the final output can be obtained through attention pooling or a single convolution layer noted by $v=W\cdot h$.

    We follow several works~\cite{gu2021vild,gao2023clip-adapter} that utilize the output features of the CLIP text encoder as the final classifier weight to align $v$ with text features $t=\mathcal{T}(S)$, where $S$ means prompt constructed by a series categories names.
    This strategy has demonstrated benefits, including the creation of a more favorable feature space for recognition~\cite{Radford:CLIP}. 
    We ensure the alignment through cosine similarity:
    \begin{equation}\label{equ:alignment strategy}
        p(y|x) = \frac{v \cdot t}{\lVert v \rVert \cdot \lVert t \rVert}.
    \end{equation}
    This method offers flexibility for expanding to unseen labels by merely adjusting the input prompt $S$ and facilitates the learning of language-aware visual representations.
    
\subsection{Combining the Discriminative prior}\label{sec:adapted expert}

    To effectively transfer learned features to discriminative tasks while ensuring compatibility, an intuitive approach is to introduce the prior knowledge of the recognition model. Leveraging the high flexibility of our U-head, we can readily fuse the diffusion prior with the discriminative prior:
    \begin{equation}\label{equ:expert}
        h = \text{U-head}([F_{SD};F_{exp}]),
    \end{equation}
    where $[ \cdot ; \cdot]$ and $F_{exp}$ means concatenation and discriminative prior respectively.

    We employ ResNet-18~\cite{resnet} to introduce discriminative prior $F_{exp}=\text{ResNet}(x)$ due to its inherent possession of multi-resolution and hierarchical semantic features. Thanks to the flexibility of our method, in fact, any discriminative model such as DINO-v2~\cite{oquab2023dinov2} can be introduced to provide guidance. More results can be found in supplementary material.

    Furthermore, to enhance the integration between discriminative prior and generative prior, we incorporated an Adapter~\cite{houlsby2019adapter} as illustrated in Figure~\ref{fig: framework}.
    Experiments demonstrate improved performance with the adapted discriminative model. Recognizing its role in enhancing compatibility with visual perception tasks, we refer to it as the ``Adapted-Expert''.

    
        

\subsection{Details of Training}\label{:train strategy}

     Initially, for the input image, we employ BLIP to obtain an accurate description of the image. Subsequently, we employ the text encoder to derive the text conditions, denoted as $c=\mathcal{T}(s)$. Following the method proposed in the previous sections, the final representations are obtained as Equation~\ref{equ:expert}. After getting the prediction according to Equation~\ref{equ:alignment strategy}, the cross-entropy loss is applied for training:
    \begin{equation}
        \mathcal{L}=\frac{1}{N} \sum^N\hat{y} \log p(y|x).
    \end{equation}

\section{Experiments}

    We verify the effectiveness of our method compared to other traditional vision models over 20+ different datasets grouped into 3 tasks, including ZS-SBIR, open-vocabulary semantic segmentation, and few-shot classification. In addition, we also show our results on a faster training schedule and validate the effectiveness of each key component in our method. 
    
\subsection{Experimental Settings}

    Unless specified, we use SD 1-5~\cite{rombach:stable-diffusion} by default and freeze all the parameters of SD model to preserve latent knowledge. For system-level comparisons, we select a few typical methods that employ different pre-training strategies, such as DINO~\cite{dino-v1} (contrastive learning), ConvNeXt~\cite{convneXt-v1}, Swin-Transformer~\cite{swin-v1} (supervised learning), MAE~\cite{he2022mae} (masked image modeling), and BeiTv3~\cite{Wang2023beitv3} (Multi-modality learning).

    \textbf{ZS-SBIR}. Following the general setting~\cite{liu2019SAKE}, we report the mean Average Precision (mAP) of our method on three datasets: Sketchy, TU-Berlin, and QuikDraw. Given the heterogeneity of the sketch and image domain, this task serves as a robustness test for the models. We train our model for 1 epoch, and the learning rate is set to 1e-4.
    
    \textbf{Open-Vocabulary Semantic Segmentation}. Following the general setting~\cite{zhou2022maskclip}, we train on the COCO-Stuff dataset and evaluate the performance on the validation set of five datasets: ADE20K-150 (ADE-150), ADE20K-847 (ADE-847), Pascal VOC (VOC), Pascal Context-59 (PC-59), and Pascal Context-459 (PC-459). Training our model for 120k iterations under weak data augmentation default and 8k iterations for the fast training schedule, we report the mean Intersection over Union (m-IoU) at a single scale. 

    \textbf{Few-shot Classification}. Following the setting of CLIP-Adapter~\cite{gao2023clip-adapter}, we report the 16-shot classification accuracy on 11 datasets: ImageNet, Caltech101 (for general object recognition), OxfordPets, StanfordCars, Flowers102, Food101, FGVCAircraft (fine-grained image recognition), EuroSAT (satellite image classification), UCF101 (action recognition), DTD (texture classification), and SUN397 (scene recognition). We train our model for 100 epochs by default and 10 epochs for a fast training schedule.

    Detailed information on training procedures and datasets can be seen in the supplementary material.

\subsection{Main Results}

    \begin{table}[t]
        \setlength{\abovecaptionskip}{0.2cm}       
        \centering
        \setlength{\tabcolsep}{0.8mm}{
        \scalebox{0.92}{
        \begin{tabular}{r|l|l|l}
             config & Classification & Segmentation & ZS-SBIR \\
             \Xhline{1.5pt}
             prompt & \multicolumn{3}{c}{BLIP}  \\
             time steps & 200 & 10 & 200\\
             attention map & \multicolumn{2}{c}{w. up cross-att. map} & w.o cross-att. map \\
             clip proj & \multicolumn{2}{c}{w.o. projection} & w. projection \\
             noise schedule & \multicolumn{2}{c}{ddpm schedule} & ddim inv \\ 
             stage in U-net & mid + down & \multicolumn{2}{c}{mid + up} \\
        \end{tabular}}
        }
        \caption{Main configuration of our method.}
        \label{tab:main configuation}
    \end{table}
    
    In this section, we present the main results and compare them with the counterparts under the default settings. The configuration across three tasks is outlined in Table~\ref{tab:main configuation}. Each row corresponds to a distinct setting in the three tasks, with detailed explanations of each configuration provided in Section~\ref{sec:sensitive analysis}.

\subsubsection{ZS-SBIR}

    ZS-SBIR is an appealing task as it closely resembles real-world scenarios. For a given sketch query $x_s$, natural images $x_i$ of the same category are retrieved based on feature similarity, which is usually measured by mAP. This task requires models to balance domain heterogeneity between sketch and image and identify unseen categories. To fairly assess the model's performance, we augment the counterpart models with an Adapter and align visual features with the text features, as mentioned in Equation~\ref{equ:alignment strategy} to facilitate knowledge transfer and mitigate the semantic gap.

    \begin{figure}[t]
        \setlength{\abovecaptionskip}{0.2cm}
        \centering
        \includegraphics[width= 0.9 \linewidth]{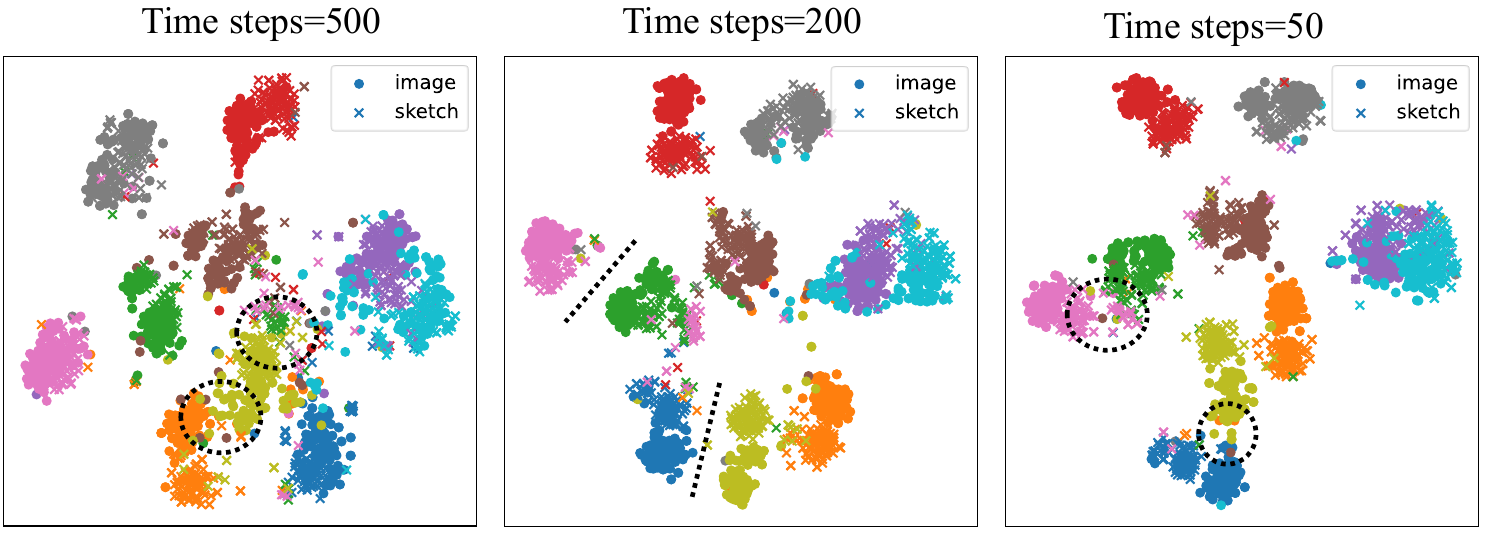}
        \caption{Feature visualization of different categories of sketches and images in different time steps.}
        \label{fig:tsne}
    \end{figure}
    
    As demonstrated in Table~\ref{tab:zs-sbir}, our method outperforms all the traditional methods, indicating that internal features in the SD model are more adept at handling abstract representations of sketches. This phenomenon is illustrated in Figure~\ref{fig:tsne}, where the disparity in abstraction levels between images and sketches diminishes as a result of light perturbation introduced at the appropriate time step (e.g., $t=200$). This leads to a more regular distribution within the feature space.
    

\begin{table}
    \setlength{\abovecaptionskip}{0.2cm}
    \centering
    \small
    \setlength{\tabcolsep}{1.5mm}{
    \begin{tabular}{l|rrr}
        \diagbox{model}{mAP} & Sketchy  & TU-Berlin & QuickDraw  \\
        \Xhline{1.5pt}
        MAE-L & 39.23 & 41.99 & 11.71   \\
        BeiTv3-G & 54.54 & 50.93 & 13.67 \\
        Swinv2-L & 43.39 & 45.51 & 12.08  \\
        DINO-B & 38.51 & 25.49 & 10.15  \\
        \hline
        Ours & \textbf{56.8} & \textbf{52.83} & \textbf{15.11}  \\
    \end{tabular}
    }
    \caption{Main results on ZS-SBIR task. Compared to traditional visual models, we achieve the best results, which is marked in \textbf{bold}}
    \label{tab:zs-sbir}
\end{table}

\subsubsection{Open-Vocabulary Semantic Segmentation}

    \begin{table}
    \setlength{\abovecaptionskip}{0.2cm}
    \centering
    \small
    \setlength{\tabcolsep}{0.5mm}{
    \begin{tabular}{l|rrrrrrr}
        \diagbox{model}{m-IoU} & \rotatebox{45}{\# param} &\rotatebox{45}{ADE-150}  & \rotatebox{45}{PC-59} & \rotatebox{45}{VOC20} & \rotatebox{45}{ADE-847} & \rotatebox{45}{PC-459} & \rotatebox{45}{\demphs{COCO}}\\
        \Xhline{1.5pt}
        MAE-L & 442 & 17.5 & 53.27 & 93.51 & 3.42 & 8.82 & \demphs{44.88}  \\
        ConvNeXt-L & 235.3 & 18.65 & \textbf{53.42} & 94.62 & 3.53 & \textbf{9.53} & \demphs{48.0} \\
        Swin-L & 234 & 18.8 & 53.37 & \textbf{94.76} & \textbf{3.8} & 9.42 &  \demphs{\textbf{48.41}}  \\
        DINO-B & 144.4 & 17.13 & 47.84 & 92.44 & 3.16 & 7.75 & \demphs{42.78}  \\
        \hline
        Ours & \textbf{5.9} & \textbf{19.0} & 52.88 & 92.87 & 3.7 & 9.0 & \demphs{46.44} \\
    \end{tabular}
    }
    \caption{Open-vocabulary semantic segmentation results on six datasets, where param means learnable parameters. We achieve comparable results while minimizing the number of tunable parameters. Results on the training set are marked in \demphs{gray}}.
    \label{tab:ov seg}
\end{table}
    
    Open-vocabulary semantic segmentation is the most challenging task, requiring the model trained on limited categories to achieve precise recognition of arbitrary or even noisy categories (e.g., ADE847 contains 847 categories, most of which are noisy).  

    For the counterpart models, we use the UperNet~\cite{xiao2018Upernet} architecture and fine-tune the entire model as the default settings. As shown in Table~\ref{tab:ov seg}, we achieve comparable or even superior results (e.g., COCO and ADE-150) compared to the tailored methods that combine specific decoder and discriminant backbone. It's noteworthy that we only have 5.9 million learnable parameters and therefore do not possess an advantage in terms of the number of learnable parameters. This is attributed to the fact that we only fine-tuned the ``\texttt{Normalization}'' layer (only 0.2 million parameters) in U-net. However, leveraging the U-head with efficient fusion capabilities and the excellent internal representation of the SD model, we still achieved a better performance against some tailored methods.

\subsubsection{Few-Shot Classification}
    
    Few-shot classification is a challenging task that demands learning from only a few samples and generalizing efficiently across multiple scenarios. For the competitor model, we fit a linear layer (i.e., linear prob) to assess its generalization ability. Due to space constraints, we only report the top-1 accuracy in the 16-shot setting.
    
    \begin{table*}
    \setlength{\abovecaptionskip}{0.2cm}
    \centering
    \small
    \setlength{\tabcolsep}{1mm}{
    \begin{tabular}{l|rrrrrrrrrrrr}
        \diagbox{model}{acc@1} & \rotatebox{45}{OxfordPets}  & \rotatebox{45}{Flowers102} & \rotatebox{45}{FGVCAircraft} & \rotatebox{45}{DTD} & \rotatebox{45}{EuroSAT} & \rotatebox{45}{StanfordCars} & \rotatebox{45}{Food101} & \rotatebox{45}{SUN397} & \rotatebox{45}{Caltech101} & \rotatebox{45}{UF101} & \rotatebox{45}{ImageNet} & \rotatebox{45}{Avg}\\
        \Xhline{1.5pt}
        MAE-L & 91.87 & 92.04 & 36.51 & 63.74 & 87.39 & 24.15 & 59.31 & 62.08 & 94.45 & 76.55 & 39.74 & 66.17 \\ 
        BeiTv3-G & \textbf{93.79} & 97.84 & 38.34 & 72.41 & 86.11 & \textbf{62.58} & 74.42 & 71.57 & 96.9 & \textbf{84.38} & \textbf{86.95} & \textbf{78.66} \\ 
        Swinv2-L& 89.65 & \textbf{99.61} & 29.13 & \textbf{73.1} & 86.9 & 37.75 & \textbf{77.41} & \textbf{72.63} & \textbf{97.01} & 81.06 & 78.84 & 74.83\\ 
        DINO-B & 89.32 & 97.82 & \textbf{48.3} & 69 & \textbf{91.15} & 57.17 & 58.5 & 62.44 & 95.57 & 76.97 & 67.66 & 73.99 \\ 
        \hline
        Ours & 66.13 & 92.35 & 42.52 & 66.62 & 88.93 & 51.05 & 45.78 & 58.09 & 95.83 & 70.49 & 55.89 & 66.74 \\
    \end{tabular}
    }
    \caption{Main results of 16-shot learning on 11 datasets. Compared to MAE, we achieve better results on some datasets and diminish the gap with professional discriminative models.}
    \label{tab:few-shot classification}
\end{table*}

    As presented in Table~\ref{tab:few-shot classification}, our method outperforms MAE on all datasets except for UCF101, OxfordPets, and SUN397. Notably, in comparison to the MAE pre-trained on IN-21K, we observed a substantial improvement of $16.15\%$ on IN-1K. While the overall performance does not exceed that of the discriminative model, comparable results are attained on specific datasets, including FGVCAircraft, EuroSAT, and Caltech101. This indicates that the inherent advantage of discriminative models in recognition tasks, due to paradigm differences with generative models, is diminishing without employing any additional techniques.
    
    An interesting observation reveals that SD models generally do not perform well on fine-grained datasets since the model may not fully understand the distinction of each class at a fine-grained level. Some failure cases are illustrated in the supplementary material.
    
    
    \begin{figure}
    \setlength{\abovecaptionskip}{0.1cm}
    	\centering
         \begin{minipage}[c]{0.85\linewidth}
    		\centering
    		\includegraphics[width=\textwidth]{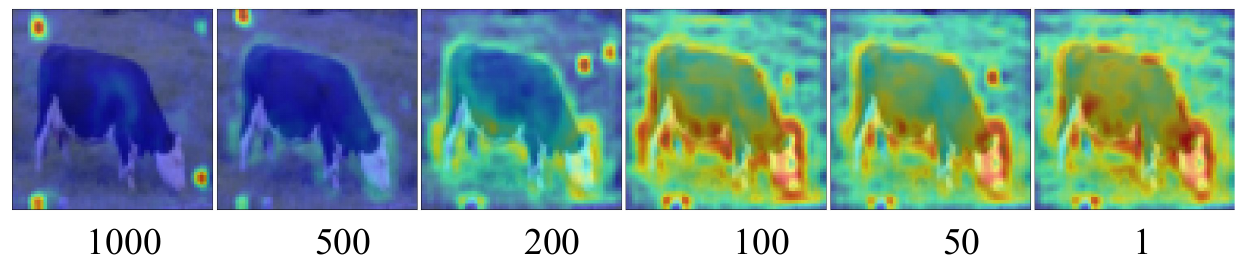}
    		\subcaption{Time step}
    		\label{fig:vis of time steps}
    	\end{minipage} \\
    	\begin{minipage}[c]{0.85\linewidth}
    		\centering
    		\includegraphics[width=\textwidth]{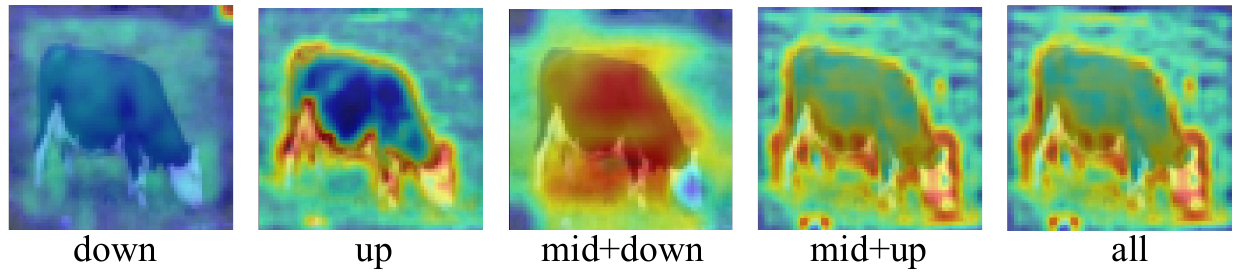}
    		\subcaption{Stage in U-net}
    		\label{fig:vis of stage in unet}
    	\end{minipage}
    	\caption{Visualization of features captured by our model at different times steps and stages of U-net}
    	\label{fig:vis of stage and time steps}
    \end{figure}


\subsection{Sensitive Analysis}\label{sec:sensitive analysis}

     In this section, we explore the potential factors affecting the performance of our method through ablation studies conducted on a faster schedule in ImageNet, Sketchy, and ADE20K (semantic segmentation). Table~\ref{tab:sensitive} reveals several intriguing properties.

\begin{table*}
    \setlength{\abovecaptionskip}{0.1cm}
    \centering
    \small
    \begin{subtable}[t]{0.45 \textwidth}
    \setlength{\abovecaptionskip}{0.05cm}
    \centering
    \setlength{\tabcolsep}{0.75mm}{
        \begin{tabular}{l|rrr}
         stage in U-net & IN-1K & Sketchy & ADE20K \\
         \Xhline{1.5pt}
         down & 36.82 & 49.8 & 33.34 \\
         up & 40.12 & 54.02 & 35.39 \\
         down + mid & \textbf{42.73} & 55.76 & 36.88 \\
         up + mid & 42.52 & \textbf{56.8} & \textbf{41.28} \\
         all & 42.04 & 53.66 & 40.53 \\
        \end{tabular}
        \caption{\textbf{Stage in U-net}. Combining mid stage with the up-sample or down-sample stage brings better results.}
        \label{tab:stage in unet}
        }
    \hfill
    \end{subtable}
    \begin{subtable}[t]{0.5 \linewidth}
    \setlength{\abovecaptionskip}{0.05cm}
    \centering
    \setlength{\tabcolsep}{0.75mm}{
        \begin{tabular}{l|rrr}
         prompt & IN-1K & Sketchy & ADE20K \\
         \Xhline{1.5pt}
         null & 29.46 & 44.71 & 39.23 \\
         random & 30.04 & 52.28 & 40.55\\
         BLIP & \textbf{42.74} & \textbf{56.8} & \textbf{41.28} \\
        \end{tabular}
        \caption{\textbf{Prompt}. An image-aligned prompt provides better guidance.}
        \label{tab:prompt analysis}
        }
    \end{subtable}

    \begin{subtable}[t]{0.3 \linewidth}
    \centering
    \setlength{\tabcolsep}{0.7mm}{
        \begin{tabular}{l|rrr}
         clip proj & IN-1K & Sketchy & ADE20K \\
         \Xhline{1.5pt}
         w.o. proj & \textbf{42.74} & 56.41 & \textbf{41.28} \\
         w. proj & 37.62 & \textbf{56.8} & 40.07 \\
        \end{tabular}
        \caption{\textbf{CLIP projection}. The text features after projected work better.}
        \label{tab:clip token}
        }
    \end{subtable}
    \hfill
    \begin{subtable}[t]{0.33 \linewidth}
    \setlength{\abovecaptionskip}{0.05cm}
    \centering
    \setlength{\tabcolsep}{0.75mm}{
        \begin{tabular}{l|rrr}
         noise & IN-1K & Sketchy & ADE20K \\
         \Xhline{1.5pt}
         w.o. & 38.59 & 47.44 & 39.87 \\
         ddim inv & 42.08 & \textbf{56.8} & 41.01 \\
         ddpm schedule & \textbf{42.74} & 54.97 & \textbf{41.28} \\
        \end{tabular}
        \caption{\textbf{Inversion method}. The results of different inversion methods have slight differences.}
        \label{tab:noise schedule}
        }
    \end{subtable}
    \hfill
    \begin{subtable}[t]{0.3 \linewidth}
    \setlength{\abovecaptionskip}{0.05cm}
    \centering
    \setlength{\tabcolsep}{0.75mm}{
        \begin{tabular}{l|rrr}
         attention & IN-1K & Sketchy & ADE20K \\
         \Xhline{1.5pt}
         w.o. & 41.32 & \textbf{56.8} & 41.03 \\
         down & 42.07 & 56.5 & 40.41 \\
         up & \textbf{42.74} & 56.57 & \textbf{41.28} \\
         up + down & 42.08 & 54.51  & 40.09 \\
        \end{tabular}
        \caption{\textbf{Cross-attention map}. Cross-attention map doesn't always seem to bring benefits.}
        \label{tab:cross attn}
        }
    \end{subtable}
    \caption{Sensitive analysis of potential factors affecting model performance in three tasks. The default setting is marked in \textbf{bold}}
    \label{tab:sensitive}
\end{table*}

\begin{figure}[t]
    \centering
    \resizebox{0.85 \linewidth}{!}{
    \includegraphics[width= \linewidth]{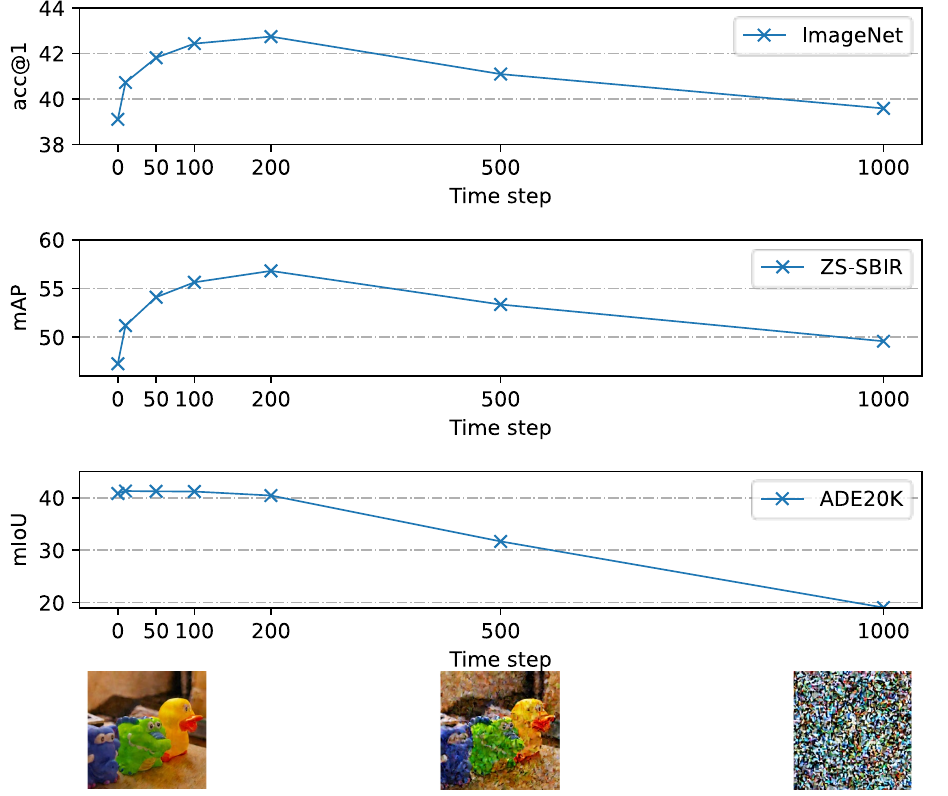}
    }
    \caption{Different time steps bring different results for three tasks.}
    \label{fig:time_steps}
\end{figure}

    \noindent\textbf{Time steps}. To validate the semantic properties of the latents under different noise-level, we evaluate our model across a series of time steps. In Figure~\ref{fig:time_steps}, we observe a consistent trend in the results of the classification and image retrieval tasks, indicating that latent representations at intermediate time steps $t \in \left[100, 200\right]$ yield better performance. However, this phenomenon is not presented in the segmentation task. This characterization is consistent with the visualization in Figure~\ref{fig:vis of time steps} regarding the analysis of different time steps. Specifically, as the images exhibit slight corruption, leading to the removal of fine-grained details intermediate, intermediate time steps focus more on capturing target subject information, whereas forward time steps exhibit a perception of the entire image. Therefore, for tasks like semantic segmentation, which demand consideration of local semantics, superior results are achieved within the time steps $t \in \left[10, 100 \right]$ as these steps preserve the capture of detailed features.
    
    \noindent\textbf{Stage in U-net}.
    As outlined in Section~\ref{sec:preliminaries}, the SD model comprises distinct stages, each encompassing different features. We selected individual stages and their associated combinations at stage granularity and carefully examined the results under various experimental configurations. As shown in Table~\ref{tab:stage in unet}, exclusively extracting features from a single stage yields suboptimal results due to the inherent incompleteness of semantics. However, when combining features from the mid-stage with either the up-sample or down-sample stage, there is a marked improvement in the outcomes. This phenomenon is consistent with Figure~\ref{fig:vis of stage in unet}, illustrating that different stages encapsulate semantic information at various granularities. Combining the mid stage with either the up-sample or down-sample stages results in more comprehensive semantic information and, consequently, superior performance.

    \noindent\textbf{Text prompt}. The key component of our method is the image-aligned prompt. In comparison to an empty prompt ($F_{SD}=\text{U-net} (z_t,t,c=\varnothing)$), image-aligned prompts can accurately reflect the content of the image and provide precise guidance. As shown in Table~\ref{tab:prompt analysis}, the prompt obtained by BLIP brought $12.7\%$, $4.52\%$, and $0.73\%$ boosts on the three tasks compared to the random prompt. We attribute this phenomenon to the fact that the image-aligned prompt is consistent with the usage of pairwise data pre-training for SD models, which allows us to retain the semantic bootstrapping abilities inherent in SD models, resulting in better performance reasonably. Contrarily, an empty or random text prompt will lead to more volatile and potentially harmful guidance.

    \noindent\textbf{CLIP token}. Some methods~\cite{zhang2023sd-dino} have demonstrated that using the second-last layer of features in ViT performs better. In the case of the CLIP model, the final layer corresponds to the \texttt{projection} layer. We are interested in exploring whether a similar phenomenon occurs in our approach, given the significant role played by the CLIP text encoder in our method. As illustrated in Table~\ref{tab:clip token}, utilizing the token from the second-to-last layer (wo. proj) leads to enhancements of $5.12\%$ and $1.21\%$ in classification and segmentation respectively. This implies that unprojected text features exhibit clearer decision boundaries for discriminative tasks and are more suitable as final classifier weights. However, performance on Sketchy has a slight decrease, which we hypothesize to be potentially associated with the diverse perceptual characteristics of different layers in the CLIP text encoder~\cite {gandelsman2023interpreting}. 

    \noindent\textbf{Inversion method}. Recently, certain approaches~\cite{mokady2023null_text_inv} have revealed that distinct inversion strategies for acquiring noisy inputs result in varied outcomes in image generation tasks. Therefore, we juxtapose two methods for introducing noise, specifically, DDIM Inversion~\cite{song2020ddim}, and the DDPM schedule~\cite{ho:ddpm}. DDIM inversion inverts latent variable $z_{0}$ into its corresponding noised version $z_{t}$ by continuousizing ordinary differential equations (ODEs). DDPM schedule adds randomness to the latent variables under the control of $\beta_{t}$.
    
    As illustrated in Table~\ref{tab:noise schedule}, the disparities between the DDPM schedule and DDIM inversion outcomes are minimal. This observation suggests that the discriminative cues within the SD model remain largely unaffected by the method of noise incorporation. However, employing the SD model to extract features from a clean image proves to be suboptimal. This inefficiency is attributed to the fact that clean latent variables do not belong to the latent space of the SD model as it is trained to predict clean images from noised version.
    
    \noindent\textbf{Cross-attention map}. Given the mechanism of injecting semantics guidance through the cross-attention in the SD model, we were prompted to investigate whether such a mechanism could also be advantageous in recognition scenarios. In this pursuit, we select the average cross-attention map from different stages in the U-net and concatenate it with our image feature, denoted as $F=\left[F_{SD}; F_{exp}; A \right]$.  
    
    As outlined in Table~\ref{tab:cross attn}, our observations indicate subtle differences in performance for the three tasks. The cross-attention map located in the up-sample stage yields an improvement of $1.42\%$ and $0.25\%$ in IN-1K and ADE20k but registers a decrease of $0.23\%$ in Sketchy. This may be caused by inaccurate attention due to the abstract nature of the sketch.  Conversely, for natural images, cross-attention maps serve as a beneficial a priori complement to the image.
    
\begin{table}
    \centering
    \small
    \setlength{\tabcolsep}{0.5mm}{
    \begin{tabular}{l|cccccc}
        \multirow{2}{*}{ } & \multicolumn{2}{c}{ZS-SBIR} & Segmentation  & \multicolumn{2}{c}{16-shot Classification} \\
        & Sketchy & Avg & ADE20K\suptext{*} & IN-1K & Avg  \\
        \Xhline{1.5pt}
        baseline & 54.28 & 40.29 & 40.3 & 37.49 & 57.32 \\
        +fuse & 55.43 & 40.99 &  40.88 & 42.38 & 59.86 \\
        +expert & \textbf{56.8} & \textbf{41.44} & \textbf{41.28} & \textbf{55.89} & \textbf{66.74} \\
    \end{tabular}
    }
    \caption{An ablation study on the key components validates the effectiveness of our approach. \suptext{*} means trained under the fast schedule}
    \label{tab:ablation study}
\end{table}

    \noindent\textbf{Discussion}. We analyzed to identify the potential factors influencing the behavior of our model. The experimental results indicate the existence of shared optimal settings applicable to both dense prediction and global recognition tasks, such as text prompts. However, certain factors, including the cross-attention and clip projection, result in inconsistent responses.

\subsection{Ablation Study}
    
    We conducted ablation studies on our two pivotal designs, U-head and Adapted-expert, following the configuration outlined in Table~\ref{tab:ablation study}. To assess the effectiveness of U-head, we established a baseline using a straightforward fusion technique, involving a convolution operation on the obtained diffusion features followed by summation. A consistent average improvement of $0.7\%, 0.58\%$, and $2.54\%$ is observed when compared to the baseline, which indicates that U-head ensures both structural unity and effectiveness. Furthermore, when combined with the Adapted-Expert, an additional improvement of $0.45\%, 0.4\%$, and $6.88\%$ is observed across all three tasks. This indicates that our method, following knowledge fusion with the discriminative model, acquires more discriminative cues, thereby enhancing discriminative accuracy. Detailed experimental results can be seen in the supplementary material. In summary, experiments across all tasks verify the effectiveness of the crucial design in our method.

\section{Conclusion}

    We introduce Vermouth, a simple yet effective unified framework that is designed to transfer the generative priors of diffusion models to discriminative tasks.  Leveraging the BLIP model, we capture an image's description as a condition, preserving the inherent advantage of semantic guidance for SD models. To accommodate various downstream tasks, we introduce a lightweight head capable of seamlessly integrating discriminative and diffuse representations within a unified framework. Experiments involving multi-tasks conducted on the unified architecture illustrate the generality and efficiency of our approach. Through the careful selection of time steps and other model components, Vermouth effectively migrates the rich visual representation of the SD mode in downstream classification, retrieval, and segmentation tasks and demonstrates a promising performance.  This exploration will not only offer valuable guidance on harnessing and optimizing the potential of SD models but also inspire further research into developing more efficient frameworks.

{\small
\bibliographystyle{ieee_fullname}
\bibliography{egbib}
}

\noindent\textbf{\large{Supplementary Material}}

\appendix
\section{Model Overview}
    
    We investigated the utilization of diffusion model, specifically focusing on the SD Model, for discriminative tasks within a unified framework. Our findings underscore that the SD Model stands out as an exemplary learner, capable of versatile application across a broad spectrum of tasks, yielding a robust representation. Despite the inherent distinctions in the training paradigms and feature spaces between generative and discriminative models, we posit the potential for leveraging synergies between these two paradigms.

    In the subsequent section, we elaborate on additional analyses of our methodology, providing a more intricate exposition of the model architecture, implementation nuances, and evaluation methodologies on three tasks, such as few-shot classification, ZS-SBIR, and open-vocabulary semantic segmentation.
    
\subsection{SD Model}
    
    SD Model comprises an encoder $\mathcal{E}$, a U-net denoted as $\epsilon_{\theta}$, and a decoder $\mathcal{D}$. Initially, the encoder maps the input to the latent space, denoted as $z_0 = \mathcal{E}(x)$, where $z_0 \in \mathbb{R}^{4 \times H \times W}$. Subsequently, the U-net engages in intricate feature acquisition and modeling of latent distributions. This involves executing conditional probability formulae to iteratively denoise noisy samples $p_{\theta}(z_{t-1}|z_t)$ to obtain clean samples $p(z_0)$, and directly model the underlying data distributions $p_{\theta}(z_{0:T})$. Once trained, a clean image can be decoded from the frozen Decoder, denoted as $\hat{x} = \mathcal{D}(z_0)$.

\begin{figure}[t]
    \centering
        \includegraphics[width= \linewidth]{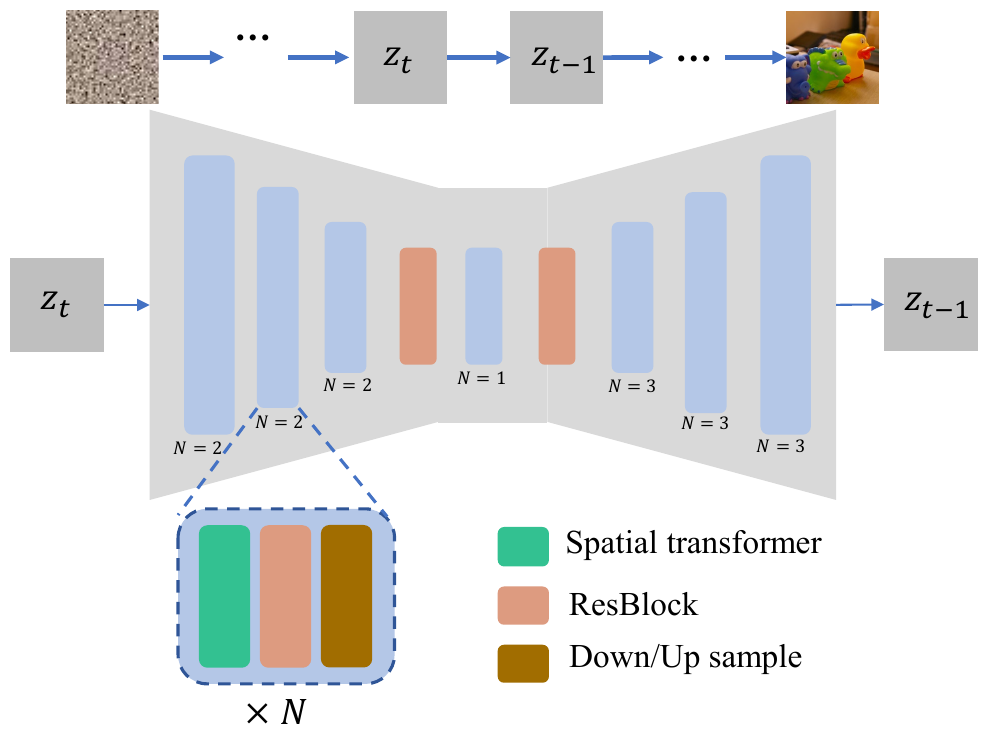}
        \caption{Architecture of U-net in the SD Model.}
        \label{fig:sd_unet}
    \end{figure}

    The pivotal aspect of our approach lies in the U-net, which furnishes a semantic-aware and robust representation of the input. Illustrated in Figure~\ref{fig:sd_unet}, the U-net comprises three modules: Spatial Transformer, ResBlock, and Down or Up Sample, constituting a total of 18 layers (the light blue block is considered as one layer). Delving into the semantic perceptual capabilities of the SD Model at the granularity of layers poses a non-trivial challenge due to the myriad of possible combinations. In this paper, we adopt a macro perspective to facilitate applicable conclusions. Our examination focuses on the difference between different stages and their combinations, utilizing the stage as the fundamental unit of analysis.

    As depicted in Figure~\ref{fig:sd_unet}, the initial three light blue blocks represent the down-sampling stage, while the concluding three light blue blocks constitute the up-sampling stage. The output for each stage is derived from the ResBlock output at different resolutions. It is essential to highlight that both the upsampling and downsampling stages incorporate features at three distinct resolutions:
    \begin{equation}
        F = \{ f_i \in \mathbb{R}^{H_i \times W_i} | Hi = Wi = \frac{64}{2^i}, i=0 \cdots 2\}.
    \end{equation}
    
    Through Classifier-free guidance, the SD Model facilitates the conditional injection of arbitrary weights, enabling the adjustment of the extent to which the output is influenced by the conditional control.
    \begin{equation}
        \hat{\epsilon}_{\theta}\left(x_{t} \mid y\right)=\epsilon_{\theta}\left(x_{t} \mid \emptyset\right)+s\cdot\left(\epsilon_{\theta}\left(x_{t} \mid y\right)-\epsilon_{\theta}\left(x_{t} \mid \emptyset\right)\right),
    \end{equation}
    where $s$ is the guidance scale coefficient (CFG).
    We believe that this ability for language perception as well as the capability for image synthesis is useful in discriminative tasks. As such, we regard the SD Model as a backbone network in an endeavor to leverage this untapped knowledge.
    
\subsection{CLIP Model}    

    We employ the CLIP ViT-L/14 text encoder to encode text into latent text embeddings, utilizing this encoded information as a condition denoted as $c = \mathcal{T}(s)$ to guide the diffusion model for extracting semantic-aware representations.
    
    In the context of aligned training, CLIP establishes a well-characterized potential feature space. Following recent works~\cite{gu2021vild,rao2022denseclip}, we recommend utilizing the text features derived from the CLIP text encoder as our final discriminant weights, represented by $W = \mathcal{T}(s) \in \mathbb{R}^{N\times d}$, where $N$ denotes the number of defined categories, and $d$ represents the feature dimension. This approach facilitates seamless expansion into unseen classes and enables efficient transfer by simply modifying the input of the text encoder to obtain $\hat{W} = \mathcal{T}(\hat{s})$, where $\hat{s}$ corresponds to the prompt involving unseen classes.

\subsection{BLIP Model}   

    The distinctive feature of BLIP resides in its integration of an encoder and decoder, culminating in the creation of a unified multi-modality model capable of comprehending and generating content. In particular, the BLIP Caption model is utilized to generate accurate descriptions of images. Operating as an image-based text decoder, the Caption model emphasizes Language Modeling (LM) throughout the training phase:
    \begin{equation}
        \max _{\theta} \mathbb{E}_{ \boldsymbol{s}_{z} \sim Z_{m}}\sum_{i=1}^{m} \log p_{\theta}\left(\boldsymbol{s}_{z_{i}} \mid \boldsymbol{s}_{\text {corrupt }}, \boldsymbol{s}_{\boldsymbol{z}_{<i}}\right),
    \end{equation}
    where $\boldsymbol{s}_{z}$ means sample from $Z_m$ text segment and $\boldsymbol{x}_{\text {corrupt }}$ means corrupted token. Upon training on a web image $I_w$, the Captioner embedded within BLIP produces synthetic captions, denoted as $T_s = \text{Cap}(I_w)$. Subsequently, it undergoes fine-tuning under the guidance of human annotators to decode text relevant to the provided image.

\begin{figure*}
        \centering
        \includegraphics[width= 0.9\linewidth]{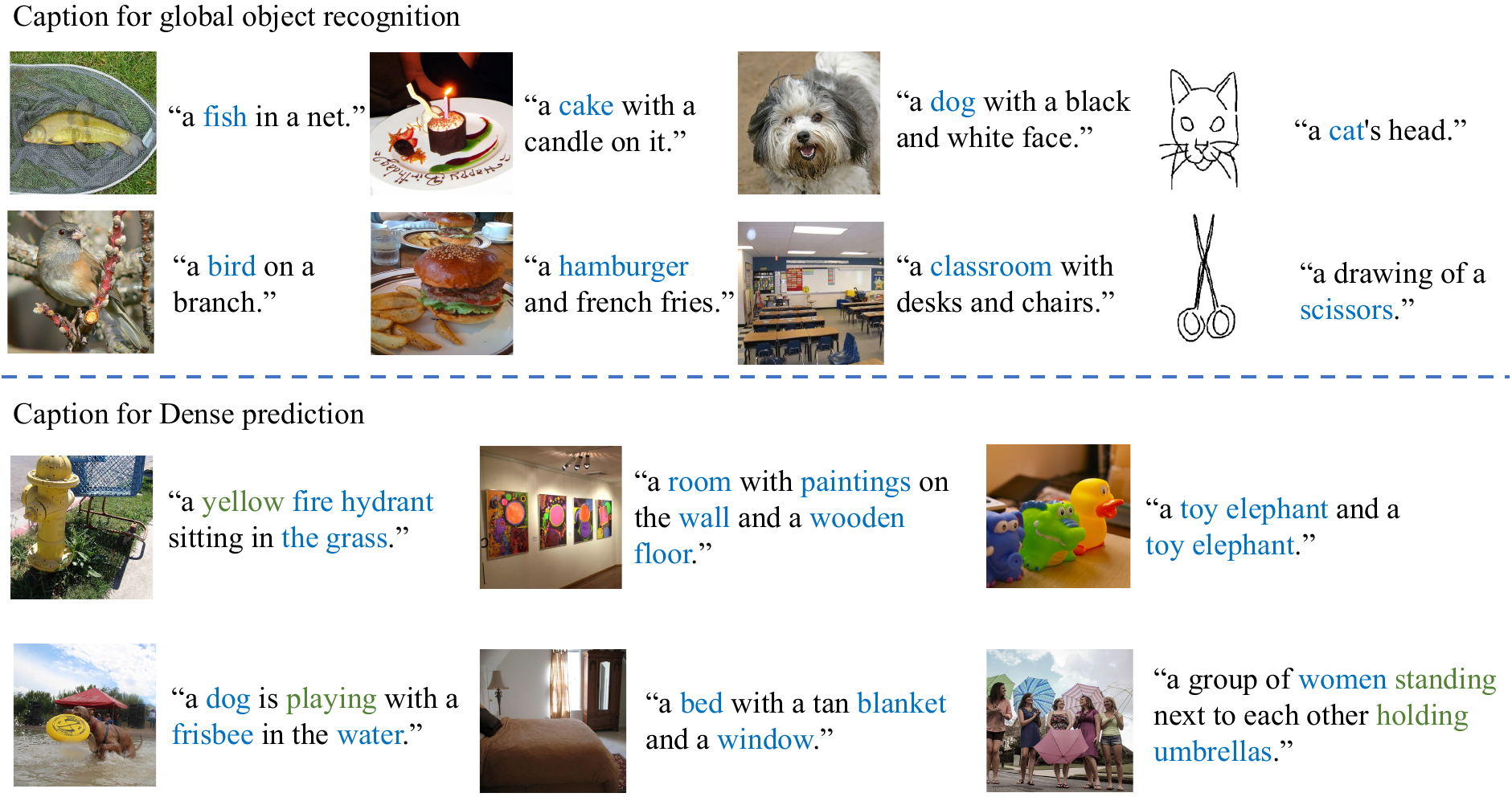}
        \caption{Caption generated by BLIP. Category labels and some potential attributes are marked by \textcolor{blue}{blue} and \textcolor{deepgreen}{green}.}
        \label{fig:BLIP case}
    \end{figure*}

    Examples illustrating the use of BLIP for text generation can be observed in Figure~\ref{fig:BLIP case}.

\subsection{U-head}

    To showcase the effectiveness of the SD Model and mitigate the risk of obtaining ambiguous conclusions resulting from the utilizing of elaborate decoder, we adhere to a straightforward design for the U-head, steering clear of intricate modules. Specifically, we implement a feature pyramid design, where each block incorporates a \texttt{Convolutional} layer, a \texttt{Group Normalization} layer, and a \texttt{SiLU} activation function:
    \begin{equation}
        f_{i+1}=\text{SiLU}(\text{GN}(\text{Conv2D}(x_{i+1}))).
    \end{equation}
    To incorporate semantics from various blocks, following each block, we acquire a representation with the same resolution as the subsequent layer through Up/Down Sampling. This process enables the capturing and aggregation of information at different granularities:
    \begin{equation}
        \widetilde{x_i} = \left[ \text{Re-Sample}(f_{i-1}); x_i\right],
    \end{equation}
    where $\left[ ; \right]$ means concatenation.

    Boasting a four-layer lightweight design, the U-head comprises a mere 8.6 million parameters, significantly smaller than a conventional FPN (46 million)~\cite{panoptic-fpn}. Despite its modest parameter count, it exhibits remarkable performance, attributed to the robust feature representation inherent in the SD Model. Furthermore, its adaptability enables the incorporation of image representations from other models, thereby enhancing the overall model representation.

    Finally, we posit that the U-head can be further enhanced through meticulous design, providing a promising avenue for future exploration.
    
\section{Concurrent Methods}

    We enumerate the concurrent methods that share similarities with our work, elucidating the distinctions from these methods along with underscoring the advantages inherent in our approach.
    
    \noindent\textbf{VPD}. Compared with VPD~\cite{zhao2023:VPD}, it has the following differences.
    \begin{itemize}
        \item Framework. VPD uses some task-specific decoders. However, experiments on three tasks with different granularity requirements demonstrate the generality of our approach using lightweight heads.
        \item Methodology. The text embedding in VPD is global and not aligned to the input. It means that the condition passed to the cross-attention layer may not appear in the input. We use the text aligned to the image to guide the model in learning the language-aware representation.
        \item Training Strategy. VPD fine-tunes the whole model. However, we explored and used parameter-efficient fine-tuning methods that did not break the original knowledge.
    \end{itemize}
    
    \noindent \textbf{Grounded-Diffusion}. Compared with Grounded-Diffusion~\cite{li2023grounded-diffusion}, we have the following advantages:
    \begin{itemize}
        \item Framework. We do not need an additional Grounding model to assist in model training. An additional model implies redundant training costs and other factors that need to be tuned.
        \item Application. Compared with the application only in the segmentation, We explored more properties of the SD Model in three tasks that cover different granularity needs.
    \end{itemize}

    \noindent \textbf{ODISE}. Compared with VPD~\cite{zhao2023:VPD}, the difference as following:
    \begin{itemize}
        \item Framework. ODISE uses the mask2former~\cite{cheng2022masked2former} as a decoder for open vocabulary segmentation and achieves the sota performance. Instead, we propose a unified architecture to explore the internal priors and application of SD Model in many downstream tasks.
        \item Methodology. ODISE uses an image captioner, which leads to potential misalignment and inefficiency problems. Specifically, In the process of injecting conditional information, cross-attention is trained for text embedding. Directly replacing them with image features may result in loss of image information as well as misalignment between the expected inputs and the actual inputs.
    \end{itemize}
    
\section{Implementation Details}

\begin{figure*}
    \centering
    \includegraphics[width=\linewidth]{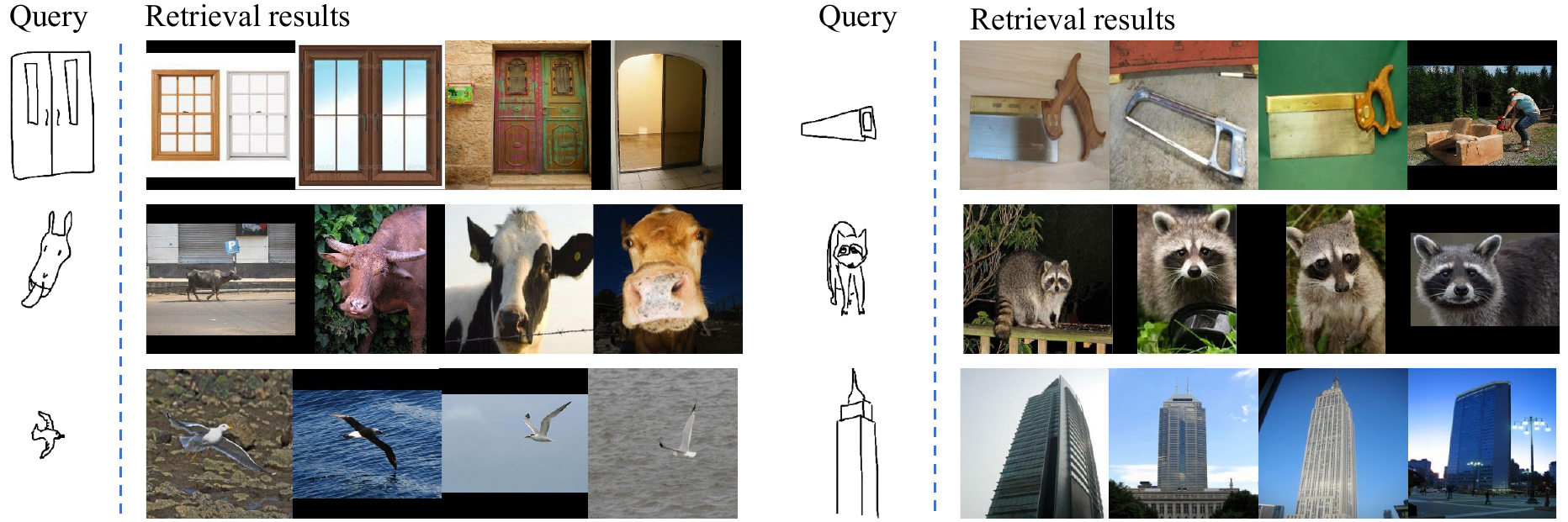}
    \caption{Zero-shot sketch-based image retrieval results. We report the model’s retrieval results on the sketch dataset}
    \label{fig:retrieval results}
\end{figure*}

\subsection*{Training}

    We trained our model following the configurations outlined in Table~\ref{tab:main configuration} for the three tasks. In the sampling process, we utilized discrete-time DDPM~\cite{ho:ddpm}, incorporated classifier-free guidance~\cite{ho:classifier-free-guidance}, and employed 1000 diffusion time steps.
    \begin{table*}[t]
        \centering
        \begin{subtable}[t]{0.35 \linewidth}
        \centering
        \setlength{\tabcolsep}{0.75mm}{
            \begin{tabular}{l|ccc}
             & Clssification & ZS-SBIR & Segmentation \\
             \Xhline{1.5pt}
             Input scale & \multicolumn{3}{c}{0.18215} \\
             CFG & \multicolumn{3}{c}{7.5} \\
             Classifier & \multicolumn{3}{c}{Text features} \\
             Guidance & \multicolumn{3}{c}{Classifier-free guidance} \\
             Noise Schedule  & \multicolumn{3}{c}{scaled linear schedule} \\
             SD Version  & \multicolumn{3}{c}{v1-5} \\ 
             BLIP & \multicolumn{3}{c}{ViT-B} \\
             U-head &  \multicolumn{3}{c}{Conv $\times$ 1, GN $\times$ 1, SiLU $\times$ 1}  \\
             Projection & \multicolumn{2}{c}{Attention pool} & Conv \\
            \end{tabular}
            \caption{\textbf{Model configuration}.}
            \label{tab:model configuration}
            }
        \end{subtable}
        \hfill
        \begin{subtable}[t]{0.55 \linewidth}
        \centering
        \setlength{\tabcolsep}{0.5mm}{
        \begin{tabular}{c|ccc}
             config & Clssification & ZS-SBIR & Segmentation \\
             \Xhline{1.5pt}
             Weight Decay &\multicolumn{3}{c}{5e-4} \\
             Optimizer &  \multicolumn{3}{c}{AdamW} \\
             Gradient Clipping Norm & \multicolumn{3}{c}{L2 norm=1} \\
             Batch Size & \multicolumn{2}{c}{64} & 2$\times$8 \\
             Temperature Coefficient & \multicolumn{2}{c}{0.2} & 0.02 \\
             Resize & \multicolumn{2}{c}{256} & 512 \\
             Learning Rate & 1e-5 & 1e-4 & 5e-4 \\
             Learning Rate Schedule  & None & Exponential & Polynomial \\
             finetune & \multicolumn{2}{c}{None} & Norm Layer \\
             \#gpus & RTX3090$\times$ 1& RTX3090$\times$ 1 & V100$\times$ 8\\
             
        \end{tabular}
        }
        \caption{\textbf{Training configuration}.}
        \label{tab:training configuation}
        \end{subtable}
        \caption{Main configuration of our method.}
        \label{tab:main configuration}
    \end{table*}

\subsection{Dataset Information}
\subsubsection{Few-Shot Classification}
    \begin{itemize}
        \item \textbf{ImageNet}. A large-scale dataset that contains over 1.2 million nature images in total. It includes diverse objects among 1000 classes in our daily lives and is a very common benchmark.
        \item \textbf{Caltech101}. A data set consisting of 101 categories of object images. It is mainly used for object recognition and image classification.
        \item \textbf{OxfordPets}. A pet image dataset containing 37 classes of pets, each with around 200 pet images. 
        \item \textbf{StanfordCars}. An image dataset of 196 car types with 16,185 images, of which 8,144 are training images and 8,041 are test images.
        \item \textbf{Flowers102}. A flower collection dataset that contains 8,189 images in total which is divided into 102 categories.
        \item \textbf{Food101}. An image dataset of 101 food categories, mainly used for fine-grained image categorization, with a total of 101,000 images. 
        \item\textbf{FGVCAircraft}. Fine-Grained Visual Classification of Aircraft is an aircraft categorization dataset, which contains 10,200 aircraft images among 102 aircraft categories.
        \item\textbf{EuroSAT}. A remote sensing image dataset that covers 13 spectral bands and consists of 10 categories. There are 27,000 images of 64 × 64 pixels in size. Each scene category contains between 2,000 and 3,000 remote sensing images.
        \item\textbf{UCF101}. An action recognition dataset with 101 action categories (e.g., Baseball Pitch, Pizza Tossing) that contains realistic action videos collected from YouTube.
        \item\textbf{DTD}. A texture recognition dataset consisting of 5,640 images was categorized into 47 classes based on human perception (e.g., striped, freckled).
        \item\textbf{SUN397}. A large-scale scene understanding and recognition dataset that contains 899 categories (e.g., abbey, castle) and 130,519 images in total.
    \end{itemize}
    We follow the CoOp~\cite{zhou2022coop} split for all datasets to ensure fairness. The split file can be found \href{https://github.com/KaiyangZhou/CoOp/blob/main/DATASETS.md}{here}

\begin{figure*}
    \centering
    \includegraphics[width=\linewidth]{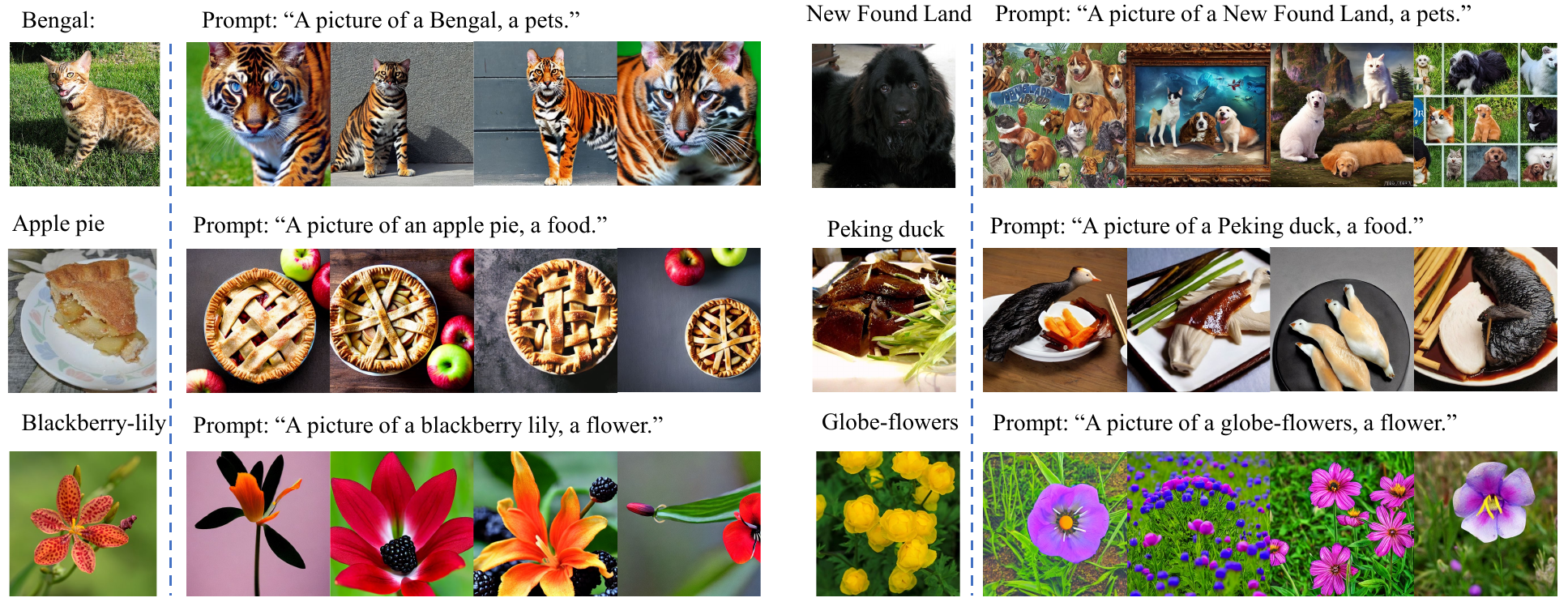}
    \caption{Some failure cases for the SD Model in comprehending fine-grained data, the ground-truth images are depicted on the left, and the synthesized images are presented on the right. For each category, we provide an appropriate prompt.}
    \label{fig:SD_failure}
    \vspace{-3mm}
\end{figure*}

\begin{table*}
    \setlength{\abovecaptionskip}{0.2cm}
    \centering
    \small
    \setlength{\tabcolsep}{1mm}{
    \begin{tabular}{l|rrrrrrrrrrrr}
        \diagbox{model}{acc@1} & \rotatebox{45}{OxfordPets}  & \rotatebox{45}{Flowers102} & \rotatebox{45}{FGVCAircraft} & \rotatebox{45}{DTD} & \rotatebox{45}{EuroSAT} & \rotatebox{45}{StanfordCars} & \rotatebox{45}{Food101} & \rotatebox{45}{SUN397} & \rotatebox{45}{Caltech101} & \rotatebox{45}{UF101} & \rotatebox{45}{ImageNet} & \rotatebox{45}{Avg} \\
        \Xhline{1.5pt}
        baseline & 37.49 & 83.49 & 31.84 & 62.76 & 85.48 & 38.53 & 42.47	& 52.21	& 94.23	& 64.55	& 37.49	& 57.32  \\
        +fuse & 38.65 & 86.25& 37.13 & 65.58 & 84.81 & 42.87 & 42.87 & 56.21 & 94.96 & 66.8 & 42.38 & 59.86 \\
        +resnet expert & {66.13}& {92.35} & \textbf{42.52} & {66.62} & {88.93} & {51.05} & {45.78} & \textbf{58.09} & {95.83} & \textbf{70.49} & \textbf{55.89} & {66.74} \\
        \hline
        +dinov1 expert & \textbf{68.67} & \textbf{96.15} & 41.89 & \textbf{69.35} & \textbf{93.32} & \textbf{59.85} & \textbf{56.46} & 55.15 & \textbf{96.48} & 70.36 & 51.4 & \textbf{69.01} \\
        +dinov2 expert & 40.82 & 87.46 & 36.79 & 65.14 & 85.75 & 44.2 & 43.07 & 56.26 & 95.08 & 67.53 & 42.17 & 60.39 \\
    \end{tabular}
    }
    \caption{Ablation results of our key designs on 11 datasets. We report accuracy@1 about 16-shot learning. The best results are marked in \textbf{bold}}
    \label{tab:supp few-shot classification}
    \vspace{-3mm}
\end{table*}

\subsubsection{ZS-SBIR}
    \begin{itemize}
        \item \textbf{Sketchy}. A large-scale dataset of fine-grained aligned sketch image
        pairs. It contains 75,471 hand-craft sketches and 73,500 nature images in 125 categories. we also adopt a 24/101 for testing and training split strategy to ensure the testing categories do not overlap with the upstream dataset.
        \item\textbf{TU-Berlin}. It contains 20k sketches over 250 categories and 204,486 natural images. Following the general splitting, we select 30 categories for testing and another 220 for training.
        \item\textbf{QucikDraw}. A novel large-scale dataset contains 110 categories containing 330,000 sketches and 204,000 images and separated 30/80 categories for testing and training. As sketches can be rough conceptual abstractions of images produced in an amateur drawing style, it has an extensive domain gap between non-expert drawers and photos.
    \end{itemize}

\subsubsection{Open Vocabulary Segmentation}
    \begin{itemize}
        \item \textbf{COCO-Stuff}. It comprises 164k images across 171 annotated classes, delineated into the training set, validation set, and test set, encompassing 118k, 5k, and 41k images, respectively. In our experiment, we use the full training set to train our model and use validation set to test.
        
        \item\textbf{ADE-150}. It encompasses a diverse collection of indoor and outdoor scene images, comprising 20k for training and 2k for validation. The dataset encapsulates a comprehensive set of 150 annotated classes.
    
        \item\textbf{ADE-847}. It has the same images as ADE-150 but more noisy annotated labels (847 classes). At the same time, because it is more in line with the open vocabulary setting, it is also the most challenging dataset.

        \item\textbf{VOC}. Pascal VOC includes annotations for semantic segmentation across 20 classes. The training set consists of 1,464 images, and the validation set contains 1,449 images.

        \item\textbf{PC-59}. Pascal Context-59 comprises 59 semantic segmentation classes and includes 5k training images along with an additional 5k validation images. 

        \item\textbf{PC-459}. It has the same images as Pascal Context-59 but more annotated classes (459 classes).
    \end{itemize}

\section{Main Results}

\begin{figure*}
    \centering
    \includegraphics[width= 0.9\linewidth]{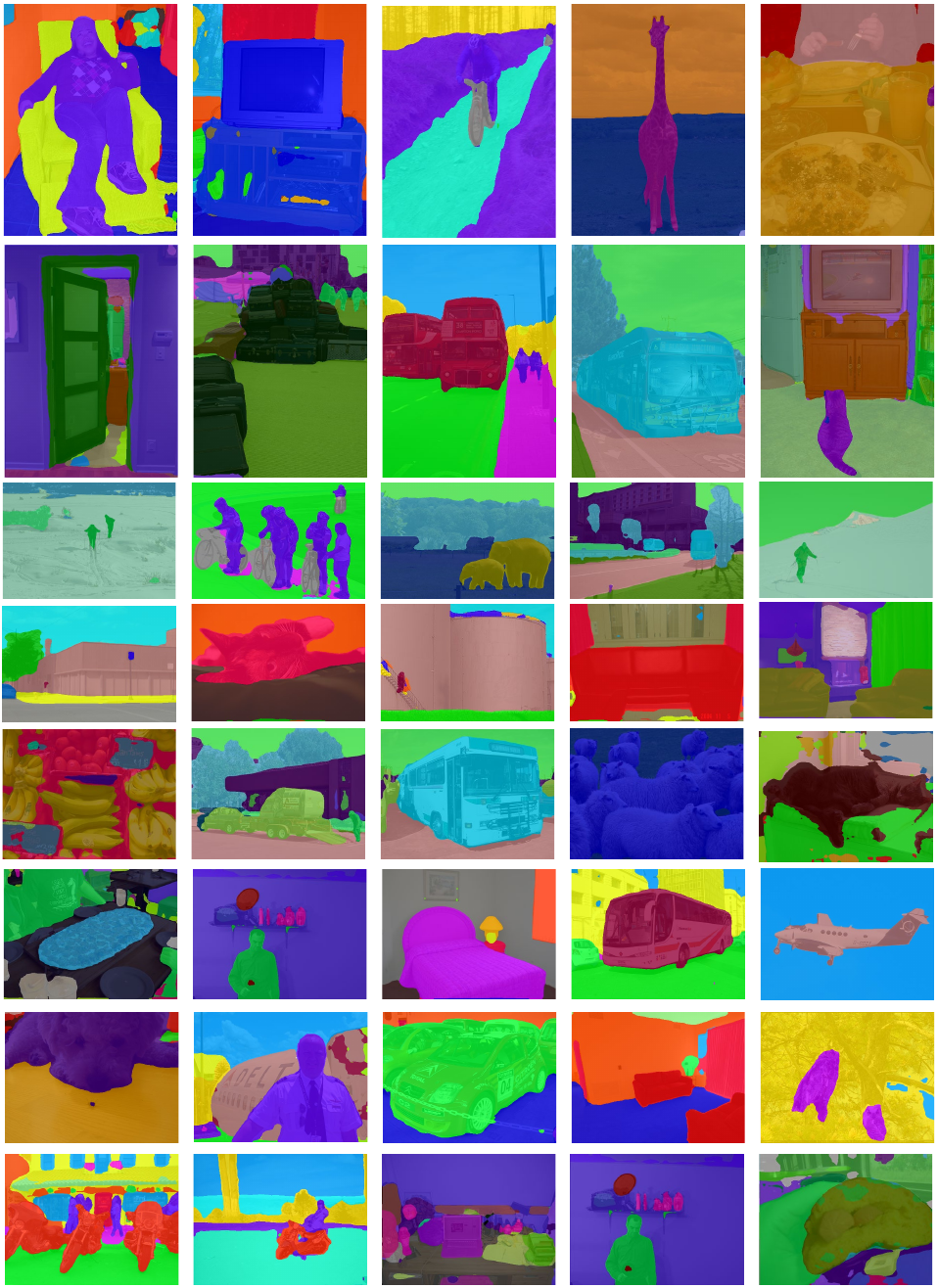}
    \caption{Qualitative visualization of open-vocabulary semantic segmentation results. We provide segmentation results on the validation set of five test datasets}
    \label{fig:seg_results}
\end{figure*}

\begin{table}
    \setlength{\abovecaptionskip}{0.1cm}
    \centering
    \small
    \setlength{\tabcolsep}{0.5mm}{
    \begin{tabular}{l|cccccc}
        \multirow{2}{*}{Expert} & \multicolumn{2}{c}{ZS-SBIR} & Segmentation  & \multicolumn{2}{c}{16-shot Classification} \\
        & Sketchy & Avg & ADE20K\suptext{*} & IN-1K & Avg  \\
        \Xhline{1.5pt}
        w.o. & 55.43 & 40.99 &  40.88 & 42.38 & 59.86 \\
        ResNet18 & \textbf{56.8} & \textbf{41.44} & \textbf{41.28} & \textbf{55.89} & 66.74 \\
        DINO-v2 & 56.43 & 41.41 & 40.38 & 42.17 & 60.39 \\
        DINO-v1 & 55.8 & 40.78 & 40.42 & 51.4 & \textbf{69.01} \\
    \end{tabular}
    }
    \caption{Discriminative Priors. We provide the performance of three different discriminative models as the Adapted Expert.}
    \label{tab:expert ablation}
    \vspace{-5mm}
\end{table}

\subsection{Baselines}
\textbf{Generative Pre-training}.
\begin{itemize}
    \item MAE~\cite{he2022mae} is a model comprised of an encoder $\mathcal{E}$ and a decoder $\mathcal{D}$. The encoder projects the input into a high-dimensional feature space, while the decoder maps these features back to the image space within an unmasked context. In our approach, we exclusively utilize the encoder to extract image features, denoted as $ f = \mathcal{E}(x)$.
\end{itemize}

\begin{figure*}
    \centering
    \includegraphics[width=0.9\linewidth]{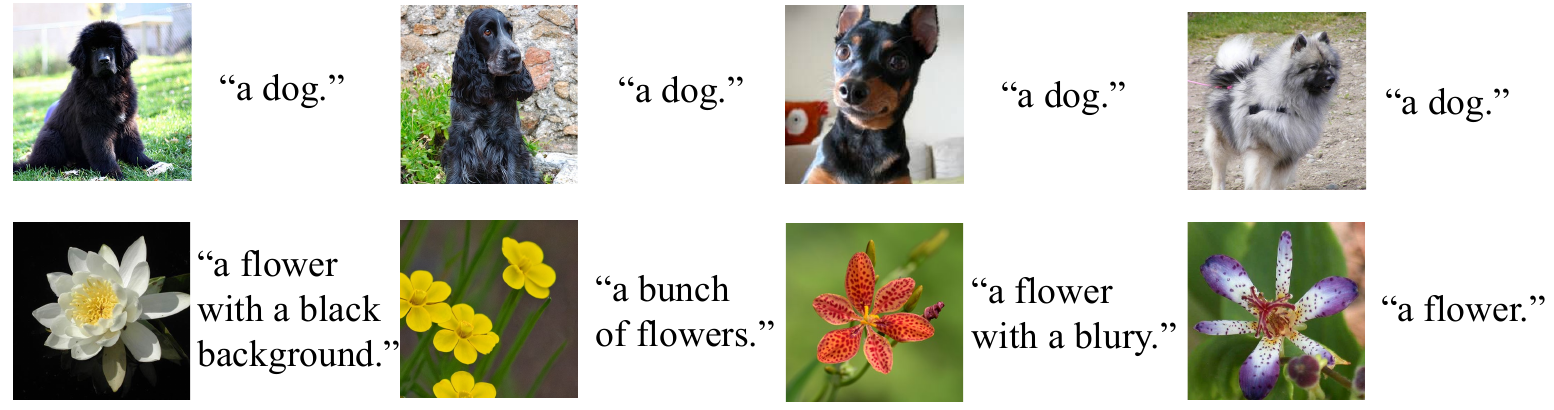}
    \caption{Some failure case about BLIP caption in some fine-grained samples. BLIP captions are positioned to the right of the images, which may be ambiguous when presented with fine-grained images.}
    \label{fig:BLIP_failure}
\end{figure*}


\noindent \textbf{Contrastive Learning}.
\begin{itemize}
    \item DINO~\cite{dino-v1} is structured with a backbone network $E$ and a projection head $H$. The backbone is employed to extract image features, denoted as $f = E(X)$. Simultaneously, the projection head maps the image representation to a higher-dimensional feature space and learns the relationships between positive and negative examples.
    
\end{itemize}

\noindent \textbf{Supervised Learning}
\begin{itemize}
    \item Swin-Transformer~\cite{swin-v1} introduces an innovative Transformer architecture that incorporates a shifted window and hierarchical structure. This design aims to reduce computational complexity while simultaneously accommodating images of various scales.
    \item ConvneXt~\cite{convneXt-v1} preserves the advantages of pure convolutional neural networks through its network structure design at both macro and micro levels. In our approach, we employ it directly as a backbone network to extract multi-resolution features.
    
\end{itemize}
    
\noindent \textbf{Multi-modality Learning}
\begin{itemize}
    \item BEiT-v3~\cite{Wang2023beitv3} adopts a MultiWay Transformer~\cite{bao2022vlmo} structure. Unlike conventional Transformer layers, this structure integrates a shared multi-head self-attention module and a Feed Forward Network (FFN) selection module customized for different modalities. Specifically, three FFN modules are assigned for processing text, image, and visual-language components individually. This configuration enhances the model's capability to capture modality-specific feature information more effectively.
    
\end{itemize}

\subsection{Evaluation \& Metrics}

    We conduct evaluations for various types of tasks and purposes, encompassing few-shot classification and ZS-SBIR to assess the quality of the learned global representations. Additionally, we assess the model's generalization of spatial representations through the open-vocabulary semantic segmentation task.

\subsection{Qualitative Analysis}

\begin{table}
    \setlength{\abovecaptionskip}{0.2cm}
    \centering
    \small
    \setlength{\tabcolsep}{1.5mm}{
    \begin{tabular}{l|rrrr}
        \diagbox{model}{mAP} & Sketchy  & TU-Berlin & QuickDraw  & Avg \\
        \Xhline{1.5pt}
        baseline & 54.28 & 52.06 & 14.52 & 40.29 \\
        +fuse & 55.43	& 52.44	& \textbf{15.11} & 40.99 \\
        +resnet expert & \textbf{56.8} & 52.83 & 
        {14.7} & \textbf{41.44} \\
        \hline
        +dinov1 expert & 55.8 & 52.16 & 14.37 & 40.78  \\
        +dinov2 expert & 56.43 & \textbf{52.85} & {14.94} & 41.41  \\
    \end{tabular}
    }
    \caption{Main results on ZS-SBIR task. Compared to traditional visual models, we achieve the best results, which are marked in \textbf{bold}}
    \label{tab:supp_zs-sbir}
    \vspace{-3mm}
\end{table}

    As depicted in Figure~\ref{fig:retrieval results} and Figure~\ref{fig:seg_results}, we showcase the qualitative results of our model. 
    In Figure~\ref{fig:retrieval results}, we achieved accurate retrieval results for sketches with varying levels of abstraction, such as raccoon and cow. It demonstrates that our method excels in retrieving natural images across different categories, even for unseen classes.

    In Figure~\ref{fig:seg_results}, our method achieves precise segmentation results across five datasets covering diverse scenarios, including indoor and outdoor environments. It is evident that, after model training, our method demonstrates commendable segmentation performance on other test sets.

\subsection{Quantitative Analysis}

    We provide ablation studies on our key designs for few-shot classification and ZS-SBIR in Table~\ref{tab:supp few-shot classification} and Table~\ref{tab:supp_zs-sbir}. We noted consistent improvements across nearly all datasets when integrating our U-head (+fuse) with the baseline. This demonstrates that, in contrast to straightforward fusion methods like feature addition (baseline), our U-head achieves a more efficient fusion of feature volumes. This substantiates the effectiveness of the U-head module. Notably, the Adapted-Expert (+expert) results in an additional improvement, indicating that combining the features extracted by the SD Model with the prior information offered by the classification model enhances the compatibility in discrimination tasks. This substantiates the effectiveness of the idea of combining the expert model to enhance compatibility for downstream tasks. We aspire for these two design concepts to serve as guiding principles for broader research endeavors.

\subsection{Discriminative Priors}

    Due to the flexibility of the U-head, our model can be combined with any discriminative model such as DINO to improve compatibility on discriminative tasks. Table~\ref{tab:supp few-shot classification}, Table~\ref{tab:zs-sbir}, and Table~\ref{tab:expert ablation} shows the results when we use the DINOv2-B/14~\cite{oquab2023dinov2} and DINOv1-B/16~\cite{dino-v1} models as the Adapter-Expert.

    In few-shot classification, as shown in Table~\ref{tab:supp few-shot classification}, we observe that combining any discriminant model will bring about steady improvements compared to the setting without Adapted Expert (+fuse). The experiment across 11 datasets shows that DINOv1 performs better compared with ResNet18 when providing discriminative priors. In ZS-SBIR, we observed that the results of the three discriminative models did not differ significantly. Moreover, we noticed more consistent improvements on ResNet18 and DINOv2-B/14 compared to DINOv1-B/16. 
    
    Generally, we opt for ResNet18 as the discriminative model to provide priors, as it consistently yields more significant benefits, as demonstrated in Table~\ref{tab:expert ablation}.

\subsection{Failure Case}

    We posit that the suboptimal performance of the SD Model on certain datasets (e.g., fine-grained Flowers102 and FGVCAircraft) can be attributed to two factors: firstly, the inherent challenges of the SD Model in synthesizing fine-grained images, and secondly, ambiguous description issues within the BLIP model when describing fine-grained data. The absence of attribute annotation can contribute to confusion in these instances.

    Figure~\ref{fig:SD_failure} depicts several instances of failure in synthesizing fine-grained data. Particularly noteworthy is the challenge faced by the SD Model in faithfully restoring the true appearance of images for specific fine-grained classes, such as Blackberry-lily and Peking duck. This limitation stems from the inherent knowledge absence within the SD Model. To address this challenge, methods like LoRa~\cite{hu2021lora} can be employed to expand the vocabulary of the SD Model, potentially enhancing its performance on fine-grained datasets.

    Another contributing factor to the undesired performance is that the BLIP Model excels at generating image-level captions, which might be overly ambitious for fine-grained datasets. As demonstrated in Figure~\ref{fig:BLIP_failure}, BLIP produces the same description ("a dog") for four distinct categories of dogs, and similarly, ambiguous descriptions are generated for different types of flowers. A potential solution is to leverage more advanced multimodal models, such as LLaVa~\cite{liu2023llava} or GPT-4~\cite{achiam2023gpt4}, to obtain more accurate captions.


\end{document}